# Text classification problems via BERT embedding method and graph convolutional neural network

John Jung
CREATORY
john@creatory.vn
Hoa Dieu Nguyen
CREATORY
hoa.nguyen@creatory.vn
Loc Hoang Tran
CREATORY
tran0398@umn.edu

Abstract: This paper presents the novel way combining the BERT embedding method and the graph convolutional neural network. This combination is employed to solve the text classification problem. Initially, we apply the BERT embedding method to the texts (in the BBC news dataset and the IMDB movie reviews dataset) in order to transform all the texts to numerical vector. Then, the graph convolutional neural network will be applied to these numerical vectors to classify these texts into their appropriate classes/labels. Experiments show that the performance of the graph convolutional neural network model is better than the performances of the combination of the BERT embedding method with classical machine learning models.

## 1. Introduction

Text classification is the very important problem in machine learning and deep learning research areas. Its applications are huge such as emotion/sentiment classification [1], intent classification [2], and genre text classification [3], etc. The outcome of this text classification problem is very valuable. Let’s discuss deeply about the advantages and the motivations of one specific text classification problem which is the genre text classification problem. After classifying one company’s products (such as video or music clips on YouTube and Facebook) into their appropriate classes/types/labels based on the descriptions of the products, we can achieve a lot of goals such as:

- Recommendation system/Analyze the match fit among users/products: This will increase the engagement of users/fans to the activities of company such as view/buy/listen to …
- Resource allocation/Optimization/Decision Making: We do know in advance which types of products that users/fans want to view/buy (and don't want to view/buy). We can put more times/human resources to produce those types of products that the users/fans want to view/buy …
- Organize all the products neatly and meaningfully. This will help the "search for products" button/function …
- We know the trends/favors of users/fans (in our network/our community). These trends/favors are very important and are very valuable.

Thus, we can easily see that text classification problem is the very valuable problem for any business. In order to solve this text classification problem, we can employ two types of approaches: the classical machine learning approach and the modern deep learning approach, to the best of our knowledge.

Initially, we need to transform the texts (the descriptions of the data samples or the comments of the users or fans) to numerical vectors. By using classical embedding approach, we can utilize the tf-idf method [4] or countvectorize method [5] or the combination of these two methods to transform the texts to numerical vectors. By using the modern embedding approach, we can utilize the BERT embedding method [6], the ELMO embedding method [7], or the FLAIR embedding method [8] to transform the texts to numerical vectors.

Finally, after obtaining the numerical vectors from texts, we can employ the classical machine learning models such as the logistic regression model [9], the random forest model [10], and the support vector machine model [11] to classify the texts into their appropriate classes/labels. Or, we can employ the modern deep learning models such as the deep convolutional neural network model [12] to classify the texts into their appropriate classes/labels.

However, there is one weakness associated with the modern deep learning models. In this case, for example, although the deep convolutional neural network model can achieve the very high performance, this model normally ignores the relationships (i.e. the connections) among the samples (i.e. texts).

In order to overcome this weakness, we propose to employ the graph convolutional neural network model [13,14,15] to solve this text classification problem. Obviously, this graph convolutional neural network model utilizes both types of information:

- The numerical vectors (for example, the BERT embedding vectors) of texts.
- The graph data structure representing the relationships among the texts.

In this paper, our contributions are three folds:

- Utilize the graph constructed from the BERT embedding vectors of the texts to solve the text classification problem.
- Employ the graph convolutional neural network model to solve the text classification problem (utilize both the BERT embedding vectors and the graph data structure representing relationships among the texts).
- Compare the performance of the graph convolutional neural network model with the performance of the combinations of the BERT embedding method with classical machine learning models.

We will organize the paper as follow: Section 2 will define the problem and will show the way how to construct the graph from the BERT embedding vectors of texts and will present the graph convolutional neural network. Section 3 will describe the two datasets that will be used in this paper which are the BBC news dataset and the IMDB movie reviews dataset and will show how to construct the graphs from the two datasets in detail and will compare the performance the graph convolutional neural network model and the performance of the combinations of the BERT embedding method and the classical machine learning models such as linear regression model testing on these two datasets. Section 4 will conclude this paper and the future directions of researches will be discussed.

## 2. Problem formulation and graph convolutional neural network

### 2.1 Problem Formulation

Given the set of the texts $\{x_1, x_2, \dots, x_l, x_{l+1}, \dots, x_{l+u}\}$ where $n = l + u$ is the total number of texts.

Please note that $\{x_1, x_2, \dots, x_l\}$ is the set of labeled texts and $\{x_{l+1}, \dots, x_{l+u}\}$ is the set of un-labeled texts.

$x_i \in R^{1*L_1}, 1 \leq i \leq n$. $L_1$ is the number of input features (i.e., the dimensions of the BERT embedding vectors). In the other words, $x_i$ is the BERT embedding vectors of the text *i*.

Let *C* be the total number of classes/labels.

Let $Y \in R^{n*C}$ be the initial label matrix. *Y* can be defined as follows

$$Y_{ij} = \begin{cases} 1\ if\ text\ i\ belongs\ to\ class\ j, 1 \leq i \leq l \\ 0\ if\ text\ i\ does\ not\ belong\ to\ class\ j, 1 \leq i \leq l \\ 0, l+1 \leq i \leq n \end{cases}$$

Our objective is to predict the labels of the un-labeled texts $x_{l+1}, \ldots, x_{l+u}$.

2.2 Construct the graph from the BERT embedding vectors

After transforming all the texts in the dataset to numerical vectors by using the BERT embedding method, we can construct the similarity graph for these text vectors by using the 5-nearest neighbor graph.

In the other words, text *i* is connected with text *j* by an edge (no direction: un-directed graph) if text *i* is among the 5 nearest neighbors of text *j* **or** text *j* is among the 5 nearest neighbor of text *i*.

We will discuss more about how to construct the similarity graph from the BERT embedding vectors in section 3.

Finally, we obtain the adjacency matrix *A* representing for the similarity graph.

$$A_{ij} = \begin{cases} 1\ if\ text\ i\ is\ connected\ to\ text\ j \\ 0\ if\ text\ i\ is\ not\ connected\ to\ text\ j \end{cases}$$

2.3 The graph convolutional neural network model

Currently, we have the set of BERT embedding vectors $\{x_1, x_2, \ldots, x_n\}$ and the adjacency matrix *A* representing the relationships among the texts in the dataset.

Please note that $x_i \in R^{1*L_1}, 1 \leq i \leq n$ and $A \in R^{n*n}$.

Let $\hat{A} = A + I$, where *I* is the identity matrix.

Let $\widehat{D}$ be the diagonal degree matrix of $\hat{A}$. In the other words, $\widehat{D}_{ii} = \sum_j \hat{A}_{ij}$.

The final output *Z* of the graph convolutional neural network can be defined as follows

$$Z = softmax(\widehat{D}^{-\frac{1}{2}}\hat{A}\widehat{D}^{-\frac{1}{2}}ReLU\left(\widehat{D}^{-\frac{1}{2}}\hat{A}\widehat{D}^{-\frac{1}{2}}X\theta_1\right)\theta_2)$$

Please note that $X$ is the input feature matrix and $X \in R^{n*L_1}$.

$\theta_1 \in R^{L_1*L_2}$ and $\theta_2 \in R^{L_2*C}$ are two parameter matrices that are needed to be learned during the training process.

*ReLU* operation can also be called Rectified Linear Unit. It is defined as follow

$$ReLU(x) = \max(0, x)$$

Normally, the *softmax* operation takes an input vector $z$ of $K$ real numbers. In the other words, $z \in R^K$ and $z = [z_1, z_2, \dots, z_K]$.

Then, the *softmax* operation can be defined as follow

$$softmax(z)_i = \frac{e^{z_i}}{\sum_j e^{z_j}}$$

For semi-supervised multiclass classification [16,17,18,19,20,21,22,23,24,25], we can then evaluate the cross-entropy error over all labeled texts:

$$L = -\sum_{k=1}^{l}\sum_{c=1}^{C} Y_{kc} lnZ_{kc}$$

The two parameter matrices $\theta_1 \in R^{L_1*L_2}$ and $\theta_2 \in R^{L_2*C}$ are trained by using the gradient descent method.

In this paper, this semi-supervised text classification problem is not like the normal supervised classification problem. In the other words, we just need a few texts with labels (for examples, we just need 10 to 100 texts with labels) [16,17,18,19,20,21,22,23,24,25]. This fact also means that we do not need thousands of labeled samples to construct the semi-supervised learning models. Then, we can apply the graph convolutional neural network to classify a couple of thousands of unlabeled texts to their appropriate classes/labels. This is the very strong argument of the semi-supervised learning framework. Last but not least, the performance of the graph convolutional neural network is at least as good as the performance of the combinations of the BERT embedding method and the classical machine learning models. This comparison will be shown in the Experiments and Results section.

3. Experiments and Results

3.1 Dataset descriptions

In this paper, we will use two datasets which are the BBC news dataset and the IMDB movie reviews dataset to test our approaches.

The BBC news dataset contains 2,126 short texts. These 2,126 short texts belong to 5 classes which are sport, business, politics, tech, and entertainment. After pre-processing these short texts, we apply the BERT embedding method to these short texts in order to transform these short texts to numerical vectors. In the other words, we will get 2,126 numerical vectors $R^{1024}$. Finally, we need to construct the similarity graph from these 2,126 numerical vectors before applying classification models to these 2,126 BERT embedding vectors (and the similarity graph).

The IMDB movie reviews dataset contains 5,000 short texts. These 5,000 short texts belong to 2 classes which are 1 (positive/good review) and 0 (negative/bad review). After pre-processing these short texts, we apply the BERT embedding method to these short texts in order to transform these short texts to numerical vectors. In the other words, we will get 5,000 numerical vectors $R^{1024}$. Finally, we need to construct the similarity graph from these 5,000 numerical vectors before applying classification models to these 5,000 BERT embedding vectors (and the similarity graph).

There are three ways to construct the similarity graph from the BERT embedding vectors such as:

- The $\varepsilon$-neighborhood graph: Connect all the texts whose pairwise distances are smaller than $\varepsilon$.
- $k$-nearest neighbor graph: Text $i$ is connected with text $j$ by an edge (no direction: un-directed graph) if text $i$ is among the $k$ nearest neighbors of text $j$ **or** text $j$ is among the $k$ nearest neighbor of text $i$.
- The fully connected graph: All texts are connected.

In this paper, the $k$-nearest neighbor graph is employed to construct the similarity graph from the BERT embedding vectors of the BBC news dataset and the IMDB movie reviews dataset. Please note that $k$ is chosen to be 5.

3.2 Experimental results

In this section, we will try to compare the accuracy performance of the graph convolutional neural network model with the accuracy performance of the combinations of the BERT embedding method with the classical machine learning models which are the logistic regression model, the random forest model, and the support vector machine model. We test our models (Python code) on Google Colab with NVIDIA Tesla K80 GPU and 12 GB RAM. The accuracy performance Q is defined as follows

$$Q = \frac{TP + TN}{TP + FP + TN + FN}$$

True Positive (TP), True Negative (TN), False Positive (FB), False Negative (FN) are defined in the following table 1:

Table 1: Definitions of TP, TN, FP, FN

| | | Predicted Label | |
|---|---|---|---|
| | | Positive | Negative |
| Known Label | Positive | True Positive (TP) | False Negative (FN) |
| | Negative | False Positive (FN) | True Negative (TN) |

For the BBC news dataset, the accuracy performances of the graph convolutional neural network model, the combinations of the BERT embedding method and the logistic regression model, the random forest model, and the support vector machine model are given in the following table 2:

Table 2: BBC news dataset: Comparison of graph convolutional neural network model with the combination of BERT embedding method with classical machine learning models. Accuracies (%) is reported.

| Number of labeled texts | 10 | 20 | 30 | 40 | 50 |
|---|---|---|---|---|---|
| Graph Convolutional Neural | **76.42** | **87.70** | **91.79** | **90.17** | **90.22** |
| Network | | | | | |
| Logistic Regression | 55.81 | 79.87 | 86.59 | 87.78 | 88.01 |
| Random forest | 35.30 | 63.20 | 79.91 | 81.59 | 82.23 |
| Support Vector Machine | 62.24 | 80.86 | 86.64 | 87.20 | 88.54 |

For the IMDB movie reviews dataset, the accuracy performances of the graph convolutional neural network model, the combinations of the BERT embedding method and the logistic regression model, the random forest model, and the support vector machine model are given in the following table 3:

Table 3: IMDB movie reviews dataset: Comparison of graph convolutional neural network model with the combination of BERT embedding method with classical machine learning models. Accuracies (%) is reported.

| Number of labeled texts | 70 | 80 | 90 | 100 | 110 |
|---|---|---|---|---|---|
| Graph Convolutional Neural Network | **73.79** | **73.49** | **73.40** | **73.95** | **74.51** |
| Logistic | 71.05 | 71.52 | 71.69 | 72.47 | 72.17 |
| Re-gres-sion | | | | | |
| Ran-dom forest | 56.77 | 63.77 | 64.42 | 63.49 | 61.39 |
| Sup-port Vector Ma-chine | 68.58 | 68.23 | 68.82 | 71.61 | 69.84 |

3.3 Discussions

From the above table 2 and table 3, we easily recognize that the graph convolutional neural network model is at least as good as the combinations of BERT embedding method with the logistic regression model, the random forest model, and the support vector machine model but often leads to better accuracy performance since graph convolutional neural network is the semi-supervised learning model. This semi-supervised learning model works perfectly even when the labeled text is scarce. However, when the number of labeled text in the training set is large (i.e. a couple of thousands labeled texts), the performances of the classical supervised learning models such as logistic regression model might be better than the performance of the semi-supervised learning model.

## 4. Conclusions

In this paper, our contributions are three folds:

- Construct the similarity graph from the BERT embedding vectors.
- Apply the graph convolutional neural network to the two text datasets to solve the text classification problem.
- Compare the performance of the graph convolutional neural network model with the performances of the combination of the BERT embedding method with the classical machine learning models.

Even though the research task constructing the similarity graph from the BERT embedding vectors looks easy, it’s the very hard task. In this paper, we just build the similarity graph from thousands of vectors. In the real cases, we need to build the similarity graph from millions or billions of vectors (i.e. representing for millions or billions of texts). The way how we do this task is really naïve and sequential. In the future, we need to figure out how to achieve this task in the smart way. For example, run our code in parallel and/or employ some specific data structure such as k-d tree. This research, to the best of our knowledge, is not only important for our business but also for government, army, etc.

In this paper, we propose and employ the graph convolutional neural network to solve the text classification problem. To the best of our knowledge, this work is not complete. There are various types of graph convolutional neural networks. In the future, we will employ the graph convolutional neural network with attention [14] and the hypergraph convolutional neural network [15] to solve this text classification problem.

**Acknowledgement**: This research was conducted with the support of research funding from CREATORY.